\documentclass[11pt]{article}

\usepackage{microtype} 
\usepackage{booktabs}  
\usepackage{url}  

\usepackage{amsmath}
\usepackage{amsthm}
\usepackage{graphicx}
\usepackage{algorithm}
\usepackage{algpseudocode}
\usepackage{amssymb}
\newcommand{\kernelname}{\texttt{Conditional Kernel}}
\newcommand{\calX}{\mathcal{X}}

\usepackage[final,workshop]{automl}
\usepackage[sort&compress,round]{natbib}
\definecolor{linkcolor}{HTML}{6929C4}
\definecolor{citecolor}{HTML}{0043CE}
\hypersetup{colorlinks={true},linkcolor=linkcolor,citecolor=citecolor}

\title{Sample-Efficient Bayesian Optimization with Transfer Learning for Heterogeneous Search Spaces}

\author[1,$\ast$]{\nameemail{Aryan Deshwal}{adeshwal@umn.edu}}
\author[2,$\ast$]{\nameemail{Sait Cakmak}{saitcakmak@meta.com}}
\author[2]{\nameemail{Yuhou Xia}{susanxia@meta.com}}
\author[2]{\nameemail{David Eriksson}{deriksson@meta.com}}

\affil[1]{University of Minnesota, Twin Cities}
\affil[2]{Meta}
\affil[$\ast$]{Equal contribution.}

\hypersetup{%
  pdfauthor={}, 
  pdftitle={},
  pdfsubject={},
  pdfkeywords={}
}

\begin{document}

\maketitle

\begin{abstract}
Bayesian optimization (BO) is a powerful approach to sample-efficient optimization of black-box functions.
However, in settings with very few function evaluations, a successful application of BO may require transferring information from historical experiments. 
These related experiments may not have exactly the same tunable parameters (search spaces), motivating the need for BO with transfer learning for heterogeneous search spaces.
In this paper, we propose two methods for this setting.
The first approach leverages a Gaussian process (GP) model with a conditional kernel to transfer information between different search spaces.
Our second approach treats the missing parameters as hyperparameters of the GP model that can be inferred jointly with the other GP hyperparameters or set to fixed values.
We show that these two methods perform well on several benchmark problems.

\end{abstract}

\section{Introduction}
Bayesian optimization (BO) is a popular technique for sample-efficient black-box optimization that has been successfully leveraged for a wide range of applications such as hyperparameter tuning for machine learning models~\citep{turner2021bayesian, snoek2012practical}, A/B testing~\citep{letham2019constrained}, chemical engineering~\citep{hernandez2017parallel}, materials science~\citep{ueno2016combo}, control systems~\citep{candelieri2018bayesian}, and drug discovery~\citep{negoescu2011knowledge}.
Many approaches to BO target the setting of expensive black-box functions where we only have access to a small number of function evaluations.
To further improve the performance of BO in this setting, transfer learning can be used as a way of leveraging relevant historical information.
While transfer learning has been thoroughly studied in the BO literature, traditional transfer learning approaches typically assume that all related experiments have exactly the same (homogeneous) search space.
This assumption simplifies the modeling process, but significantly limits the applicability of these methods in real-world scenarios where search spaces often vary between experiments. 
For instance, when tuning the hyperparameters of machine learning models, practitioners often make changes to the search space to add and/or remove parameters or update their ranges.

In this regime, we need a BO method that can seamlessly transfer information from several historical experiments with different search spaces.
To achieve this, we propose two methods targeting the setting of a small number of function evaluations where we do not have access to additional domain knowledge, e.g., user priors.
Our first method leverages a conditional kernel that defines a similarity measure between heterogeneous spaces by leveraging a tree-structured representation of the search spaces. 
This method has the advantage that it requires no additional hyperparameters.
Our second method employs a learned imputation which treats missing parameters as hyperparameters that can be learned jointly with the other GP hyperparameters. 
Both methods allow incorporating task-level similarity information and naturally correspond to standard transfer learning BO in the special case when the search spaces are identical. Our main contributions are:
\begin{enumerate}
    \item We propose two methods (conditional kernel-based and learned imputation-based) for BO with transfer learning  for heterogeneous search spaces.
    \item We empirically validate our methods on benchmark datasets, demonstrating better sample-efficiency compared to existing approaches.
    \item We provide a BoTorch implementations of both approaches.
\end{enumerate}

\section{Related Work}
\paragraph{Transfer Learning in Bayesian Optimization (BO)} 
BO has been previously explored with the common goal of leveraging information from historical \emph{source} tasks to improve optimization efficiency on a new \emph{target} task~\citep{bai2023transfer}. 
Early work by~\citep{swersky2013multi,yogatama2014efficient}) employed multi-task Gaussian Processes (MTGPs) to model task similarities.
In addition,~\citep{tighineanu2022transfer} provide a unified view of hierarchical GP models for transfer learning in BO. 
Ensemble methods have also been utilized as surrogate models in the context of transfer learning~\citep{schilling2016scalable,feurer2018scalable}. 

Recent work has also considered learning neural network-parameterized GP priors from previous tasks~\citep{perrone2018scalable,wistuba2021few,wang2021pre}. 
Additionally, \citep{shilton2017regret, wang2018regret, dai2022provably} performed a theoretical analysis for the regret metric commonly employed in BO.
However, all these approaches focus on the setting where the search spaces are homogeneous across tasks, i.e., all tasks share the same search space.
This limits these methods applicability for problems with different (heterogeneous) search spaces.

\paragraph{BO with Heterogeneous Search Spaces}
\citep{fan2024transfer} recently proposed a method for transfer learning across heterogeneous search spaces that leverages a neural network mapping from domain-specific contexts to specifications of hierarchical GPs. 
The data requirements of neural networks and the need for manually defined domain-specific contexts may limit the applicability of this method in settings where small amounts of data is available.
Another class of recent methods leverage text-based sequential modeling approaches for meta black-box optimization~\citep{chen2022towards, song2024omnipred}.
While this sequential modeling can naturally handle heterogeneous search spaces, these methods require massive amounts of training data. 
During paper review, we became aware of \citep{stoll2020hyperparameter}, which motivates the heterogeneity through the adjustments made across a sequence of hyperparameter optimizations for machine learning models. They introduce the problem, a series of benchmark problems and baseline algorithms; and demonstrate that transferring knowledge across experiments can lead to significant savings in experimentation budget. \footnote{We thank the reviewer who pointed this reference to us during the review period.}

\section{Background}
\label{sec:background}

\paragraph{Bayesian Optimization}
Bayesian optimization (BO) is an iterative approach to black-box optimization, see~\citep{frazier2018tutorial,garnett_bayesoptbook_2023} for a comprehensive overview. 
BO consists of two main steps where we first build a probabilistic surrogate model from available data followed by optimizing an acquisition function to select the most promising candidate(s) to evaluate next.
This iterative process continues until the evaluation budget has been exhausted.
The probabilistic surrogate model is commonly a Gaussian process (GP)~\citep{rasmussen2006gaussian} and a common choice of acquisition function is the expected improvement (EI)~\citep{jones98}.

\paragraph{Bayesian Optimization with Transfer Learning}
Our goal is to optimize a {\em target} black-box function (task) $f_t(x)$ where each evaluation of $f_t(x)$ is expensive and the number of function evaluations is limited. 
We assume a budget of $n$ function evaluations for $f_t(x)$, where $n$ is commonly between $5$ and $40$.
In the {\em transfer learning} setting we are also given data from a set of $t-1$ related optimization experiments, referred to as {\em source} tasks, $f_1, f_2, \ldots, f_{t-1}$.
We want to leverage this existing data to improve the sample-efficiency of BO on the target task $f_t(x)$.  

In this paper, we are interested in the setting where tasks are {\em heterogeneous}, i.e., correspond to different search spaces.
As an example, consider tuning a few hyperparameters of a simple neural network with source tasks $f_1(x), f_2(x)$ and target task $f_3(x)$ with search spaces defined as follows:
\begin{itemize}
    \item Search space of $f_1$: $\calX_{1}$ =  \{learning rate, dropout rate\}
    \item Search space of $f_2$ : $\calX_{2}$ =  \{learning rate, dropout rate, batch size\}
    \item Search space of $f_3$: $\calX_{3}$ =  \{learning rate, dropout rate, number of hidden layers\}
\end{itemize}
In this simple example, the common parameters are \{learning rate, dropout ratio\}, with the search spaces of $f_{2}(x)$ and $f_{3}(x)$ having some additional parameters not present in the other tasks.

\section{BO with Transfer Learning for Heterogeneous Search Spaces}
In this section, we will propose two methods for BO with transfer learning for heterogeneous search spaces.
Our first method uses a conditional kernel which allows us to leverage a GP model to correlate the common parameters across tasks.
Our second method treats the missing parameters for each task as hyperparameters that will need to be inferred when training the GP model.
A clear benefit of these methods compared to existing methods in the literature, e.g.,~\citep{fan2024transfer,chen2022towards, song2024omnipred}, is that neither method assumes additional information beyond the previously evaluated inputs and corresponding function values of the source and target tasks.

\subsection{MTGP with Conditional Kernels} 
\label{sec:conditional_kernel}
Next, we will describe a GP model that employs a new kernel (referred as \kernelname) to handle the challenge of modeling heterogeneous search spaces across different tasks. 
The key idea behind \kernelname{} is to leverage a set of base kernels, e.g., Matern-5/2, to {\emph conditionally} compare inputs from different tasks based on their matching parameters in a dependency tree representing the search spaces of all the tasks.

\begin{figure}[!ht]
    \begin{center}
        \includegraphics[width=0.5\textwidth]{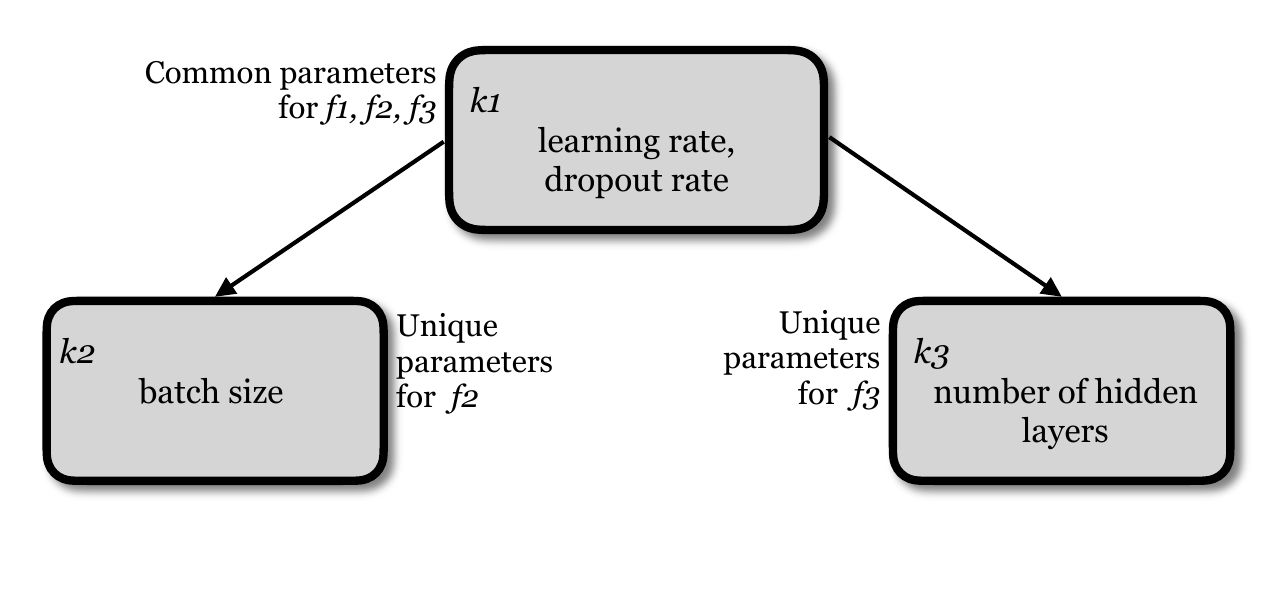}
        \caption{The tree-based representation and the corresponding kernels of the search spaces $\calX_{1}, \calX_{2}$ and $\calX_{3}$ given in Sec.~\ref{sec:background}.}
        \label{fig:kernel_viz}
    \end{center}
\end{figure}

\paragraph{Tree-Structured Representation of Heterogeneous Search Spaces} 
To make this more concrete, let $\mathcal{U} = \bigcup_{i=1}^t \calX_{i}$ be the universal set of parameters from all the tasks. 
We create a tree-representation of this universal set $\mathcal{U}$ where each node of the tree corresponds to a subset of parameters. 
Starting from the root node, we assign a maximal subset of parameters to each node that are common across as many tasks as possible. 
Subsequently, we define a set of base kernels that is equal to the number of nodes in the tree. 
Fig.~\ref{fig:kernel_viz} illustrates this tree for the example given in Sec.~\ref{sec:background}.  

\vspace{-5ex}

\paragraph{Conditional kernel} 
The \kernelname{} ($\mathcal{K}_c$) of any two inputs $x$ and $x'$ coming from two tasks $i, i'$ is defined as the sum of base kernels over all nodes that contain the common parameters across both task $i$ and task $i'$: 
\begin{align*}
    \mathcal{K}_c (\{x, i\}, \{x', i'\}) = \sum_{j=1}^p \textbf{\text{\LARGE{I}}}[N_{\mathcal{U}}[j] \subset \calX_{i} \cap  \calX_{i'}] \cdot k_j (x[N_{\mathcal{U}}[j]],  x'[N_{\mathcal{U}}[j]]),
\end{align*}    
where $N_{\mathcal{U}}[j]$ refers to the subset of parameters that are contained in node indexed by $j$ and $\textbf{\text{\LARGE{I}}}$ is the indicator function that determines whether the corresponding subset $N_{\mathcal{U}}[j]$ is part of tasks $\calX_1$ and $\calX_2$.
See Sec.~\ref{sec:example kernel} in the appendix for a concrete example of constructing the conditional kernel.

\paragraph{MTGP with Conditional Kernels} 
Many MTGP models leverage a base kernel for the input space and a task kernel representing the task correlations. 
In the setting of heterogeneous search spaces, we can combine the \kernelname{} $\mathcal{K}_c (\{x, i\}, \{x', i'\})$ with a parameterized positive definite matrix $B$ representing the task correlations to define the popular intrinsic model of co-regionalization (ICM) kernel~\citep{swersky2013multi}:
\begin{align*}
   \text{Overall MTGP Kernel | } k(\{x, i\}, \{x', i'\}) = \mathcal{K}_c (\{x, i\}, \{x', i'\}) \cdot B
\end{align*}

\subsection{MTGP with Imputed Values}
\label{sec:imputed_mtgp}
In many applications, the missing parameters represent fixed settings that have not been tuned in the experiment.
With this motivation, 
we propose an approach that treats each missing (from the union of search spaces) parameter in a task as an additional hyperparameter (for the unknown fixed value) in the MTGP model.
These missing parameters can either be learned jointly with the other GP hyperparameters or set to some fixed value.
While this approach introduces additional hyperparameters (one per missing parameter per task), the computational complexity of GP training is on the same order as an MTGP defined over the union of search spaces.

\section{Experiments}
\label{sec:experiment}
In this section, we provide an empirical evaluation of different methods for BO with heterogeneous search spaces.
All experiments use $100$ replications and we show the mean performance with two standard errors in all plots.
All methods use a squared exponential (SE) kernel. We use the Log-Normal priors proposed in~\citep{hvarfner2024vanilla}, which are designed to be performant in both low and high-dimensional settings, using length-scale priors that scale by $\sqrt{D}$.
We use the recently proposed LogEI extension of the popular Expected Improvement (EI) acquisition function~\citep{ament2023unexpected}. 
For each replication, we initialize all methods with exactly the same initial random points and generate a new set of random trials for the source tasks.

\paragraph{Methods} 
We compare six different methods.
\emph{Random Search} which samples randomly from the search space. 
\emph{Vanilla BO} uses standard BO on the target task and ignores the data from the source tasks.
\emph{MTGP with Conditional Kernels} uses the approach described in Sec.~\ref{sec:conditional_kernel}.
\emph{MTGP on Common Parameters} uses an MTGP + LogEI on the common parameters and samples other parameters randomly.
\emph{Imputed MTGP} uses the imputed model as described in Sec.~\ref{sec:imputed_mtgp} and sets missing parameters to the center of the parameter range.
\emph{Learnable Imputed MTGP} uses the imputed MTGP from Sec.~\ref{sec:imputed_mtgp} and learns the missing parameters jointly with the other GP model hyperparameters.

\begin{figure}[!ht]
    \begin{center}
    \includegraphics[width=0.94\textwidth]{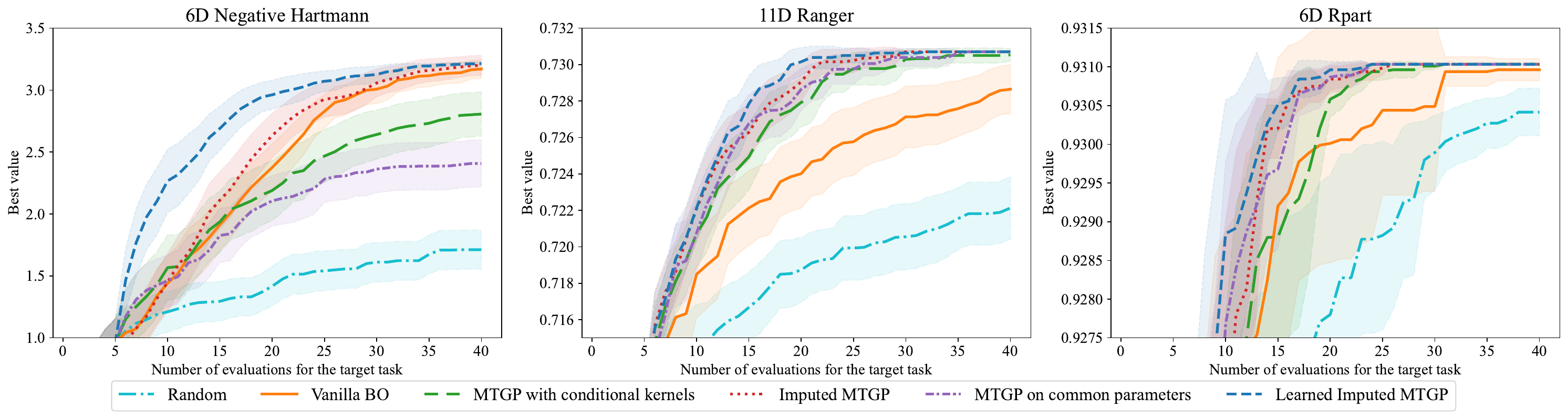}
    \caption{
        (Left) The Learned Imputed MTGP performs the best on the 6D Hartmann problem. 
        (Middle/Right) The MTGP-based methods outperform Vanilla BO and Random Search on the 11D Ranger and 6D Rpart problems. Each experiment employs 30 source task trials.
    }
    \label{fig:botl_results}
    \end{center}
\end{figure}

\paragraph{Synthetic problems}
Our first test problem uses the popular synthetic Hartmann6 function. 
We use one source task which corresponds to a 4D search space where last $2$ parameters are fixed to $0$. 
The target task is the original Hartmann6 problem.
The results are show in the left plot of Fig.~\ref{fig:botl_results}.

\paragraph{HPO-B benchmark problems}
HPO-B is a large transfer learning benchmarking suite for hyperparameter optimization~\citep{pineda2021HPOB}\footnote{The code for the HPO-B benchmark is available at \url{https://github.com/releaunifreiburg/HPO-B}}. 
We construct two different test problems based on problems available in HPO-B.
The first problem is \emph{Ranger}, which has $11$ tunable parameters. 
We use search space ids $5965$ (missing parameter index 10) and $7607$ (missing parameter index $5$ and $9$) as the source tasks.
The target task is search space id $5889$ with parameters [$0$, $1$, $2$, $3$, $6$, $7$].
The second problem is \emph{Rpart} where we use search space ids $5636$ and $5859$ (both with all $6$ parameters) as source tasks. 
The target task has search space id $4796$ and only has parameters [$0$, $2$, $3$]. 

We use $30$ source trials from each source task, resulting in a total of $60$ source trials. 
The results for the Ranger and Rpart problems are shown in the middle and right plots in Fig.~\ref{fig:botl_results}.

\paragraph{Code} All methods were implemented in Python using BoTorch \citep{balandat2020botorch}. The source code and the instructions to reproduce the benchmark results can be found in our Github repository:  \url{https://github.com/facebookresearch/heterogeneous_botl}.

\subsection{Ablation study}
\begin{figure}[!ht]
    \begin{center}
    \includegraphics[width=0.94\textwidth]{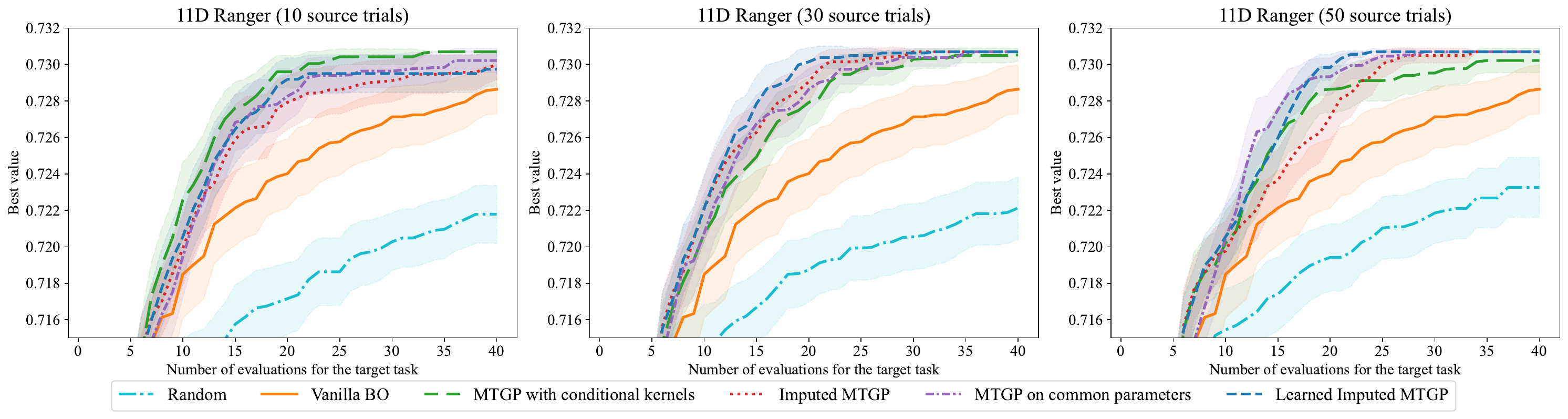}
    \caption{Ablation study for varying the number of trials for each source task on the 11D Ranger problem. 
    We observe that the MTGP with conditional kernels performs the best when only a small number of source trials are available.
    As more source data is available, the Learned Imputed MTGP and MTGP on common parameters perform the best. 
    }
    \label{fig:ranger_ablation}
    \end{center}
\end{figure}

We perform an ablation study where we vary the number of source trials on the 11D Ranger problem.
Fig.~\ref{fig:ranger_ablation} shows the results for $10$, $30$, and $50$ source trials for each source task.

\vspace{-2ex}

\section{Conclusions}
Our results show that BO with transfer learning can be successfully leveraged in the setting of heterogeneous search spaces where we have access to small amounts of historical data.
For future work, we plan on exploring ways of 
leveraging additional domain-specific information.
Additionally, leveraging other probabilistic models, e.g., Bayesian neural networks, may address some of the scaling limitation of our approach that come from leveraging exact MTGP models.

\section{Broader Impact}
After careful reflection, the authors have determined that this work presents no notable negative impacts to society or the environment. 

\bibliography{main}

\newpage
\section*{Submission Checklist}

\begin{enumerate}
\item For all authors\dots
  \begin{enumerate}
  \item Do the main claims made in the abstract and introduction accurately
    reflect the paper's contributions and scope?
    \answerYes{The abstract and introduction focus on Bayesian optimization with transfer learning which is the setting considered in the paper.}
  \item Did you describe the limitations of your work?
    \answerYes{We point out limitations of our work in the conclusion.}
  \item Did you discuss any potential negative societal impacts of your work?
    \answerYes{We do not foresee any negative societal impact of this work since this is purely methods work. We also added the discussion in the main paper.}
  \item Did you read the ethics review guidelines and ensure that your paper
    conforms to them? \url{https://2022.automl.cc/ethics-accessibility/}
    \answerYes{We confirm that we have reviewed the guidelines that our paper confirms to them}
  \end{enumerate}
\item If you ran experiments\dots
  \begin{enumerate}
  \item Did you use the same evaluation protocol for all methods being compared (e.g., same benchmarks, data (sub)sets, available resources)?
    \answerYes{All methods use the same experimental setup. We initialize all methods with exactly the same initial seed for each replication. }
  \item Did you specify all the necessary details of your evaluation (e.g., data splits, pre-processing, search spaces, hyperparameter tuning)?
    \answerYes{We describe all necessary details in the experimental section. Furthermore, we have also provided the code and will open-source the implementation upon acceptance.}
  \item Did you repeat your experiments (e.g., across multiple random seeds or splits) to account for the impact of randomness in your methods or data?
    \answerYes{We use $100$ replications for all experiments in this paper.}
  \item Did you report the uncertainty of your results (e.g., the variance across random seeds or splits)?
    \answerYes{All our figures show the mean performance +/- 2 standard errors.}
  \item Did you report the statistical significance of your results?
    \answerYes{All experiments use $100$ replications and we plot the mean +/- 2 standard errors.}
  \item Did you use tabular or surrogate benchmarks for in-depth evaluations?
    \answerYes{We consider two benchmark problems from HPO-B dataset.}
  \item Did you compare performance over time and describe how you selected the maximum duration?
    \answerNA{We report performance as a function of the number of evaluations, which is common for Bayesian optimization.}
  \item Did you include the total amount of compute and the type of resources
    used (e.g., type of \textsc{gpu}s, internal cluster, or cloud provider)?
    \answerYes{We spent a total of 6.5 CPU days of a 4-core 12 GB machine to run the experiments that were used to generate the plots in this paper. 
    We spent a negligible amount of additional compute on exploration.}
  \item Did you run ablation studies to assess the impact of different
    components of your approach?
    \answerYes{We conduct an ablation study on the number of source trials to better understand how that affects performance.}
  \end{enumerate}
\item With respect to the code used to obtain your results\dots
  \begin{enumerate}
\item Did you include the code, data, and instructions needed to reproduce the
    main experimental results, including all requirements (e.g.,
    \texttt{requirements.txt} with explicit versions), random seeds, an instructive
    \texttt{README} with installation, and execution commands (either in the
    supplemental material or as a \textsc{url})?
    \answerYes{We have included a link to an anonymized repository with our code and instructions to reproduce the results, see \url{https://anon-github.automl.cc/r/heterogeneous_botl-CD0D/}}
  \item Did you include a minimal example to replicate results on a small subset
    of the experiments or on toy data?
    \answerYes{We included full details of how to replicate the results by running the benchmarks in the README section of the code repository.}
  \item Did you ensure sufficient code quality and documentation so that someone else can execute and understand your code?
    \answerYes{We made sure the README contains proper documentation of the code including description of formatting and testing.}
  \item Did you include the raw results of running your experiments with the given code, data, and instructions?
    \answerYes{We added all the raw results in the results directory of our code repository available at the anonymous link: \url{https://anon-github.automl.cc/r/heterogeneous_botl-CD0D/} }
  \item Did you include the code, additional data, and instructions needed to generate the figures and tables in your paper based on the raw results?
    \answerYes{We provided the code to generate the figures from raw results.}
  \end{enumerate}
\item If you used existing assets (e.g., code, data, models)\dots
  \begin{enumerate}
  \item Did you cite the creators of used assets?
    \answerYes{We have cited HPO-B which we use for two benchmark problems. We have also cited the BoTorch paper since we leverage BoTorch both for our methods and the baselines.}
  \item Did you discuss whether and how consent was obtained from people whose
    data you're using/curating if the license requires it?
    \answerNA{Both BoTorch and HPO-B are freely provided without restriction, under respective MIT licenses.}
  \item Did you discuss whether the data you are using/curating contains
    personally identifiable information or offensive content?
    \answerNA{}
  \end{enumerate}
\item If you created/released new assets (e.g., code, data, models)\dots
  \begin{enumerate}
    \item Did you mention the license of the new assets (e.g., as part of your code submission)?
    \answerNo{We left license out of the submission to preserve anonymity during review. All code and assets will be released under MIT license upon publication.}
    \item Did you include the new assets either in the supplemental material or as
    a \textsc{url} (to, e.g., GitHub or Hugging Face)?
    \answerYes{Implementation of all models and benchmarking related code is provided via an anoymized GitHub link.}
  \end{enumerate}
\item If you used crowdsourcing or conducted research with human subjects\dots
  \begin{enumerate}
  \item Did you include the full text of instructions given to participants and
    screenshots, if applicable?
    \answerNA{}
  \item Did you describe any potential participant risks, with links to
    Institutional Review Board (\textsc{irb}) approvals, if applicable?
    \answerNA{}
  \item Did you include the estimated hourly wage paid to participants and the
    total amount spent on participant compensation?
    \answerNA{}
  \end{enumerate}
\item If you included theoretical results\dots
  \begin{enumerate}
  \item Did you state the full set of assumptions of all theoretical results?
    \answerNA{}
  \item Did you include complete proofs of all theoretical results?
    \answerNA{}
  \end{enumerate}
\end{enumerate}

\newpage
\appendix

\section{Example conditional kernel construction}
\label{sec:example kernel}
Without loss of generality, we can assign each parameter a unique integer id.
For the example from Sec.~\ref{sec:background}:
\begin{itemize}
    \item Search space of $f_1$: $\calX_1$ =  \{1: learning rate, 2: dropout rate\}
    \item Search space of $f_2$ : $\calX_2$ =  \{1: learning rate, 2: dropout rate 3: batch size\}
    \item Search space of $f_3$: $\calX_3$ =  \{1: learning rate, 2: dropout rate, 4: number of hidden layers\}
\end{itemize}

This means that the universal set of parameters is $\mathcal{U} = \{1, 2, 3, 4\}$.
The conditional kernel can be constructed as follows: 
\begin{enumerate}
    \item First, we construct a set  $\widetilde{\mathcal{U}}$ of size $p$ consisting of subsets of ids that are common across as many tasks as possible. 
    For example, in the case above, $\widetilde{\mathcal{U}} = \{[1, 2], [3], [4]\}$ with $p=3$. Algorithm \ref{alg:find_subsets} provides the pseudo-code for this procedure.
    
    \item We define $p$ base kernels $\{k_1, k_2, \cdots, k_p\}$ for each subset in $\widetilde{\mathcal{U}}$. 
    For example, in the case above, we define three base kernels $\{k_1, k_2, k_3\}$.
    \item The conditional kernel $\mathcal{K}_c$ value of two inputs $x$ and $x'$ coming from two tasks $i, i'$ is then defined as the sum of the base kernels that correspond to the subsets containing common  parameters of both task $i$ or task $i'$:
    \begin{align}
        \mathcal{K}_c (\{x, i\}, \{x', i'\}) = \sum_{j=1}^p \textbf{\text{\LARGE{I}}}[\widetilde{\mathcal{U}}[j] \subset \calX_{i} \cap \subset \calX_{i'}] \cdot k_j (x[\widetilde{\mathcal{U}}[j]],  x'[\widetilde{\mathcal{U}}[j]]),
    \end{align}    
    where $\textbf{\text{\LARGE{I}}}$ is the indicator function that determines if the corresponding subset $\widetilde{\mathcal{U}}[j]$ is part of both tasks $\calX_1$ and $\calX_2$ 
    We let $\widetilde{\mathcal{U}}[j]$ refer to the $j$th subset of $\widetilde{\mathcal{U}}$.  
    For example, in the example above, the conditional kernel for inputs $x$ from task 1 and $x'$ from task 2 is defined as:
    \begin{align}
        \mathcal{K}_c (\{x, i\}, \{x', i'\}) = k_1 (x[1, 2], x'[1, 2]).
    \end{align}    
    Similarly, if the inputs $x$ and $x'$ are from the same task 2, the conditional kernel is:
    \begin{align}
    \mathcal{K}_c (\{x, i\}, \{x', i'\}) = k_1 (x[1, 2], x'[1, 2]) + k_2 (x[3], x'[3])
    \end{align}   
\end{enumerate}

\begin{algorithm}
\caption{Construct $\widetilde{\mathcal{U}}$: Subsets of Common Indices}
\label{alg:find_subsets}
\begin{algorithmic}[1]
\Require List of parameter index lists $\mathcal{F} = [F_1, F_2, ..., F_t]$ where $F_i$ corresponds to task $i$
\State $\text{$\widetilde{\mathcal{U}}$} \gets \{\text{set}(F_1)\}$
\For{$i \gets 2$ to $n$}
    \State $\text{idx\_set} \gets \text{set}(F_i)$
    \State $\text{new\_subsets} \gets []$
    \For{$\text{sub} \in \text{$\widetilde{\mathcal{U}}$}$}
        \State $\text{common} \gets \text{idx\_set} \cap \text{sub}$
        \State $\text{remaining} \gets \text{sub} \setminus \text{common}$
        \If{$\text{common} \neq \emptyset$}
            \State $\text{new\_subsets}.\text{append}(\text{common})$
            \State $\text{idx\_set} \gets \text{idx\_set} \setminus \text{common}$
        \EndIf
        \If{$\text{remaining} \neq \emptyset$}
            \State $\text{new\_subsets}.\text{append}(\text{remaining})$
        \EndIf
    \EndFor
    \If{$\text{idx\_set} \neq \emptyset$}
        \State $\text{new\_subsets}.\text{append}(\text{idx\_set})$
    \EndIf
    \State $\text{$\widetilde{\mathcal{U}}$} \gets \text{new\_subsets}$
\EndFor
\State \Return $\widetilde{\mathcal{U}}$
\end{algorithmic}
\end{algorithm}

\paragraph{Remark} Our approach only transfers information between the shared parameters in any two tasks. If there is no overlap, no information can be transferred as this will result in two independent sub-kernels being trained and utilized for the two tasks.

\end{document}